\documentclass[10pt,twocolumn,letterpaper]{article}

\usepackage{cvpr}
\usepackage{times}
\usepackage{epsfig}
\usepackage{graphicx}
\usepackage{float}
\floatstyle{plaintop}
\restylefloat{table}
\usepackage{caption}
\usepackage{subcaption}
\usepackage{amsmath}
\usepackage{amssymb}
\usepackage{bm}
\usepackage{dsfont}
\usepackage{subcaption}
\usepackage{blindtext}
\usepackage{colortbl,booktabs}
\usepackage{multirow}
\usepackage{booktabs}
\usepackage[english]{babel}
\usepackage[utf8x]{inputenc}
\usepackage{algorithm}
\usepackage{algpseudocode} 
\newcommand{\minus}{\scalebox{0.75}[1.0]{$-$}}

\usepackage[pagebackref=true,breaklinks=true,letterpaper=true,colorlinks,bookmarks=false]{hyperref}

\cvprfinalcopy %

\def\cvprPaperID{6895} %

\ifcvprfinal\fi
\begin{document}

\title{Conditional Gaussian Distribution Learning for Open Set Recognition}

\author{Xin Sun$^1$, ~~Zhenning Yang$^1$, ~~Chi Zhang$^1$, ~~Guohao Peng$^1$, ~~Keck-Voon Ling$^2$\\
Nanyang Technological University, Singapore\\
{\tt\small $^1$\{xin001,zhenning002,chi007,peng0086\}@e.ntu.edu.sg, ~$^2$ekvling@ntu.edu.sg}
}

\maketitle
\thispagestyle{empty}

\begin{abstract}
   Deep neural networks have achieved state-of-the-art performance in a wide range of recognition/classification tasks. However, when applying deep learning to real-world applications, there are still multiple challenges. A typical challenge is that unknown samples may be fed into the system during the testing phase and traditional deep neural networks will wrongly recognize the unknown sample as one of the known classes. Open set recognition is a potential solution to overcome this problem, where the open set classifier should have the ability to reject unknown samples as well as maintain high classification accuracy on known classes. The variational auto-encoder (VAE) is a popular model to detect unknowns, but it cannot provide discriminative representations for known classification. In this paper, we propose a novel method, Conditional Gaussian Distribution Learning (CGDL), for open set recognition. In addition to detecting unknown samples, this method can also classify known samples by forcing different latent features to approximate different Gaussian models. Meanwhile, to avoid information hidden in the input vanishing in the middle layers, we also adopt the probabilistic ladder architecture to extract high-level abstract features. Experiments on several standard image datasets reveal that the proposed method significantly outperforms the baseline method and achieves new state-of-the-art results. 
\end{abstract}

\section{Introduction}

\begin{figure}
\begin{subfigure}{\columnwidth}
  \centering
  \includegraphics[scale=0.35]{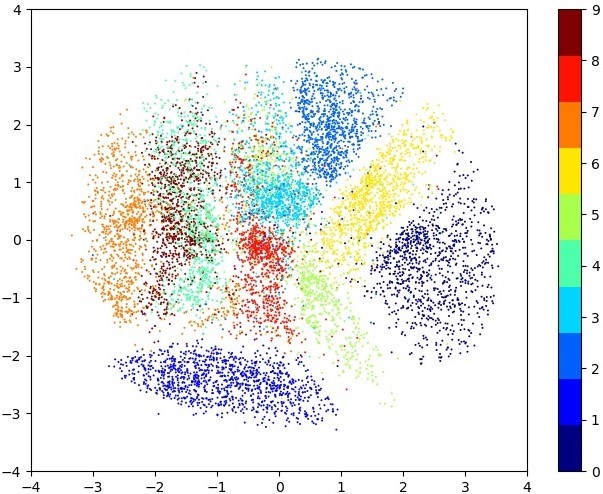}
  \caption{VAE}
  \label{fig:vae}
\end{subfigure}%
\hspace{.2in}
\begin{subfigure}{\columnwidth}
  \centering
  \includegraphics[scale=0.35]{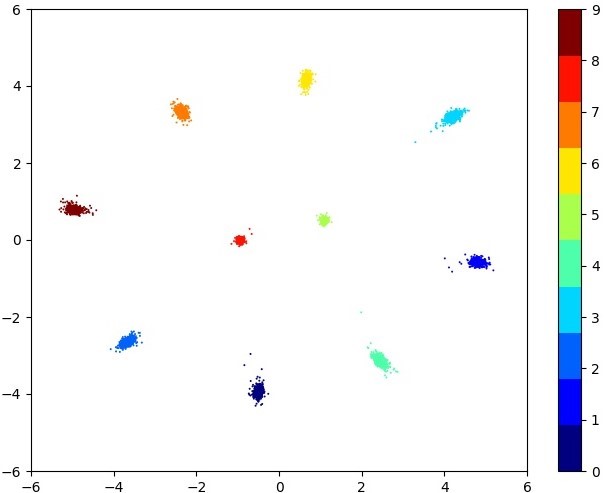}
  \caption{Ours: CGDL}
  \label{fig:cvga}
\end{subfigure}
\caption{Comparison of latent representations on MNIST dataset of the VAE (a) and the proposed method CGDL (b). The VAE is widely used in unknown detection, but it cannot provide discriminative features to undertake classification tasks as all features just follow one distribution. Comparatively, the proposed method can learn conditional Gaussian distributions by forcing different latent features to approximate different Gaussian models, which enables the proposed method to classify known samples as well as reject unknown samples.}
\end{figure}

In the past few years, deep learning has achieved state-of-the-art performance in many recognition/classification tasks \cite{mr-cnn,res-net,ssd,r-cnn}, but there are still multiple challenges when applying deep learning to real-world problems. One typical challenge is that incomplete knowledge exists during the training phase, and unknown samples may be fed into the system during the testing phase. While traditional recognition/classification tasks are under a common closed set assumption: all training and testing data come from the same label space. When meeting an unknown sample, traditional deep neural networks (DNNs) will wrongly recognize it as one of the known classes. 

The concept of open set recognition (OSR) \cite{toward} was proposed, assuming the testing samples can come from any classes, even unknown during the training phase. The open set classifier should have a dual character: unknown detection and known classification$\footnote{We refer to detection of unknown samples as \emph{unknown detection}, and classification of known samples as \emph{known classification}.}$. Considering that during training it is not available to extract information from unknown samples, to realize unknown detection, many previous works analyze information from known samples by unsupervised learning \cite{ad_cluster,deepsvdd,ad_aec,gmm}. Among them, the variational auto-encoder (VAE) \cite{vae} is a popular method, in combination with clustering \cite{ad_cluster}, GMM \cite{gmm}, or one-class \cite{deepsvdd} algorithms. The VAE is a probabilistic graphical model which is trained not only to reconstruct the input accurately, but also to force the posterior distribution $q_{\bm{\phi}}(\bm{z}|\bm{x})$ in the latent space to approximate one prior distribution $p_{\bm{\theta}}(\bm{z})$, such as the multivariate Gaussian or Bernoulli distribution. The well-trained VAE is able to correctly describe known data, and deviated samples will be recognized as unknown. Fig.~\ref{fig:vae} is an example of the VAE latent representations on MNIST dataset when the prior distribution $p_{\bm{\theta}}(\bm{z})$ is the multivariate Gaussian. Although the VAE excels at unknown detection, it cannot provide discriminative representations to undertake classification tasks as all features only follow one distribution.

Here, to overcome this shortcoming, we propose a novel method, Conditional Gaussian Distribution Learning (CGDL), for open set recognition. Different from traditional VAEs, the proposed method is able to generate class conditional posterior distributions $q_{\bm{\phi}}(\bm{z}|\bm{x},k)$ in the latent space where $k$ is the index of known classes. These conditional distributions are forced to approximate different multivariate Gaussian models $p_{\bm{\theta}}^{(k)}(\bm{z})={\mathcal{N}}(\bm{z};\bm{\mu}_{k},\textbf{I})$ where $\bm{\mu}_k$ is the mean of the $k$-th multivariate Gaussian distribution, obtained by a fully-connected layer that maps the one-hot encoding of the input's label to the latent space. Fig.~\ref{fig:cvga} is an example of latent representations of the proposed method on MNIST dataset. These learned features will be fed to an open set classifier, which consists of two parts: an unknown detector and a closed set classifier. As known samples tend to follow the prior distributions, the unknown detector will recognize those samples locating in lower probability regions as unknown. Meanwhile, for the known sample, the closed set classifier will calculate its prediction scores over all known classes and predict it as the class with the highest score.

Current networks tend to go deeper for higher accuracy in recognition/classification tasks \cite{deeper}. However, traditional VAEs are restricted to shallow models as details of input could be lost in higher layers \cite{lost}, which limits VAE's ability to extract high-level abstract features. To fully exploit information from known samples, we adopt the probabilistic ladder network \cite{ladder} into the proposed method. This probabilistic ladder architecture allows information interactions between the upward path and the downward path, which enables the decoder to recover details discarded by the encoder. Although there are several successful applications of the probabilistic ladder network \cite{sequen,infer,lost}, this paper is the first to apply it to open set recognition.

In our experiments, we explore the importance of the probabilistic ladder architecture and the conditional distributions in the latent space for open set recognition. We empirically demonstrate that our method significantly outperforms baseline methods. In summary, this paper makes the following contributions:
\begin{itemize}
\setlength
  \item We propose a novel open set recognition method, called Conditional Gaussian Distribution Learning (CGDL). Compared with previous methods based on VAEs, the proposed method is able to learn conditional distributions for known classification and unknown detection.
  \item We develop a fully-connected layer to get the means of different multivariate Gaussian models, which enables posterior distributions in the latent space to approximate different Gaussian models.
  \item We adopt a probabilistic ladder architecture to learn high-level abstract latent representations to further improve open set classification scores.
 \item We conduct experiments on several standard image datasets, and the results show that our method outperforms existing methods and achieves new state-of-the-art performance.
\end{itemize}

\section{Related Work}
\textbf{Open Set Recognition.}  The methods for open set recognition (OSR) can be broadly divided into two branches: traditional methods (e.g., SVM, sparse representation, Nearest Neighbor,etc.) and deep learning-based methods. In traditional methods, Scheirer~\emph{et~al.}~\cite{toward} proposed an SVM based method which adds an extra hyper-line to detect unknown samples. Jain~\emph{et~al.}~\cite{toward-pi} proposed the $P_I$-SVM algorithm, which is able to reject unknown samples by adopting EVT to model the positive training samples at the decision boundary. Cevikalp~\emph{et~al.}~\cite{fafr,poly} defined the acceptance regions for known samples with a family of quasi-linear `polyhedral conic' functions. Zhang~\emph{et~al.}~\cite{sparse} pointed out that discriminative information is mostly hidden in the reconstruction error distributions, and proposed the sparse representation-based OSR model, called SROSR. Bendale~\emph{et~al.}~\cite{openworld} recognized unknown samples based on the distance between the testing samples and the centroids of the known classes. J{\'u}nior~\emph{et~al.}~\cite{nearest} proposed the Nearest Neighbor Distance Ratio (NNDR) technique, which carries out OSR according to the similarity score between the two most similar classes. Considering deep learning achieves state-of-the-art performance in a wide range of recognition/classification tasks, deep learning-based open set recognition methods are gaining more and more attention.

In deep learning-based methods, Bendale~\emph{et~al.}~\cite{openmax} proposed the Openmax function to replace the Softmax function in CNNs. In this method, the probability distribution of Softmax is redistributed to get the class probability of unknown samples. Based on Openmax, Ge~\emph{et~al.}~\cite{g-openmax} proposed the Generative Openmax method, using generative models to synthesize unknown samples to train the network. Shu~\emph{et~al.}~\cite{doc} proposed the Deep Open Classifier (DOC) model, which replaces the Softmax layer with a 1-vs-rest layer containing sigmoid functions. Counterfactual image generation, a dataset augmentation technique proposed by Neal~\emph{et~al.}~\cite{counter}, aims to synthesize unknown-class images. Then the decision boundaries between unknown and known classes can be converged from these known-like but actually unknown sample sets. Yoshihashi~\emph{et~al.}~\cite{crosr} proposed the CROSR model, which combines the supervised learned prediction and unsupervised reconstructive latent representation to redistribute the probability distribution. Oza and Patel~\cite{c2ae} trained a class conditional auto-encoder (C2AE) to get the decision boundary from the reconstruction errors by extreme value theory (EVT). The training phase of C2AE is divided into two steps (closed-set training and open-set training), and a batch of samples need to be selected from training data to generate non-match reconstruction errors. This is difficult in practice and testing results are highly dependent on the selected samples. On the contrary, the proposed method is an end-to-end system and does not need extra data pre-processing.

\textbf{Anomaly Detection.} Anomaly detection (also called outlier detection) aims to distinguish anomalous samples from normal samples, which can be introduced into OSR for unknown detection.  Some general anomaly detection methods are based on Support Vector Machine (SVM) \cite{svdd,ocsvm} or forests \cite{forest}. In recent years, deep neural networks have also been used in anomaly detection, mainly based on auto-encoders trained in an unsupervised manner \cite{ad_aec, ad_cluster, deepsvdd, gmm}. Auto-encoders commonly have a bottleneck architecture to induce the network to learn abstract latent representations. Meanwhile, these networks are typically trained by minimizing reconstruction errors. In anomaly detection, the training samples commonly come from the same distribution, thus the well-trained auto-encoders could extract the common latent representations from the normal samples and reconstruct them correctly, while anomalous samples do not contain these common latent representations and could not be reconstructed correctly. Although VAEs are widely applied in anomaly detection, it cannot provide discriminative features for classification tasks.

Apart from auto-encoders, some studies used Generative Adversarial Networks (GANs) to detect anomalies \cite{ad_gan}. GANs are trained to generate similar samples according to the training samples. Given a testing sample, the GAN tries to find the point in the generator's latent space that can generate a sample closest to the input. Intuitively, the well-trained GAN could give good representations for normal samples and terrible representations for anomalies.

There are also some related tasks focusing on novel classes. For example, few-shot learning~\cite{zhang2020deepemd,zhang2019pyramid,zhang2019canet} aims to undertake vision tasks on new classes with scarce training data. Incremental learning~\cite{liu2020mnemonics} aims to make predictions on both old classes and new classes without accessing data in old classes.
\section{Preliminaries}

Before introducing the proposed method, we briefly introduce the terminology and notation of VAE \cite{vae}. 

The VAE commonly consists of an encoder, a decoder and a loss function ${\mathcal{L}}(\bm{\theta};\bm{\phi};\bm{x})$. The encoder is a neural network that has parameters $\bm{\phi}$. Its input is a sample $\bm{x}$ and its output is a hidden representation $\bm{z}$. The decoder is another neural network with parameters $\theta$. Its input is the representation $\bm{z}$ and it outputs the probability distribution of the sample. The loss function in the VAE is defined as follows:

\begin{equation}
\begin{split}
{\mathcal{L}}(\bm{\theta};\bm{\phi};\bm{x})= & \ \minus D_{KL}\big(q_{\bm{\phi}}(\bm{z}|\bm{x}) \ || \ p_{\bm{\theta}}(\bm{z})\big)\\
& +\mathds{E}_{q_{\bm{\phi}}(\bm{z}|\bm{x})}\big[\log{p_{\bm{\theta}}(\bm{x}|\bm{z})}\big]
\label{loss_vae}
\end{split}
\end{equation}
where $q_{\bm{\phi}}(\bm{z}|\bm{x})$ is the approximate posterior, $p_{\bm{\theta}}(\bm{z})$ is the prior distribution of the latent representation $\bm{z}$ and $p_{\bm{\theta}}(\bm{x}|\bm{z})$ is the likelihood of the input $\bm{x}$ given latent representation $\bm{z}$. On the right-hand side of Eqn.~\ref{loss_vae}, the first term is the KL-divergence between the approximate posterior and the prior. It can be viewed as a regularizer to encounter the approximate posterior to be close to the prior $p_{\bm{\theta}}(\bm{z})$. The second term can be viewed as the reconstruction errors.

Commonly, the prior over the latent representation $\bm{z}$ is the centered isotropic multivariate Gaussian $p_{\bm{\theta}}(\bm{z})={\mathcal{N}}(\bm{z};\bm{0},\textbf{I})$. In this case, the variational approximate posterior could be a multivariate Gaussian with a diagonal covariance structure:

\begin{equation}
q_{\bm{\phi}}(\bm{z}|\bm{x})={\mathcal{N}}(\bm{z};\bm{\mu},\bm{\sigma}^2\textbf{I})
\end{equation}
where the mean $\bm{\mu}$ and the standard deviation $\bm{\sigma}$ of the approximate posterior are outputs of the encoding multi-layered perceptrons (MLPs). The latent representation $\bm{z}$ is defined as $\bm{z}=\bm{\mu}+\bm{\sigma}\odot \bm{\epsilon}$ where $\bm{\epsilon}\sim {\mathcal{N}}(\bm{0},\textbf{I})$ and $\odot$ is the element-wise product. Let $J$ be the dimensionality of $\bm{z}$, then the KL-divergence can be calculated \cite{vae}:

\begin{equation}
\begin{aligned}
&\minus D_{KL}\big(q_{\bm{\phi}}(\bm{z}|\bm{x}\big) \ || \ p_{\bm{\theta}}(\bm{z}))\\
&=\frac{1}{2}\sum_{j=1}^{J}\big(1+\log(\sigma_{j}^{2})-\mu_{j}^{2}-\sigma_{j}^{2}\big)
\label{kl_vae}
\end{aligned}
\end{equation}

With loss function ${\mathcal{L}}(\bm{\theta};\bm{\phi};\bm{x})$, the VAE is trained not only to reconstruct the input accurately, but also to force the posterior distribution $q_{\bm{\phi}}(\bm{z}|\bm{x})$ in the latent space  to approximate the prior distribution $p_{\bm{\theta}}(\bm{z})$. If a sample locates in the low probability region of the learned distribution, this sample will be recognized as unknown.

\section{Proposed Method} 

In this section, we describe the proposed method in detail. Firstly, we describe the architecture of the proposed model. Then, we introduce the training phase and the testing phase to describe the functions of each module.
\subsection{Architecture}
\begin{figure} [htbp]
    \centering
    \includegraphics[width=\columnwidth]{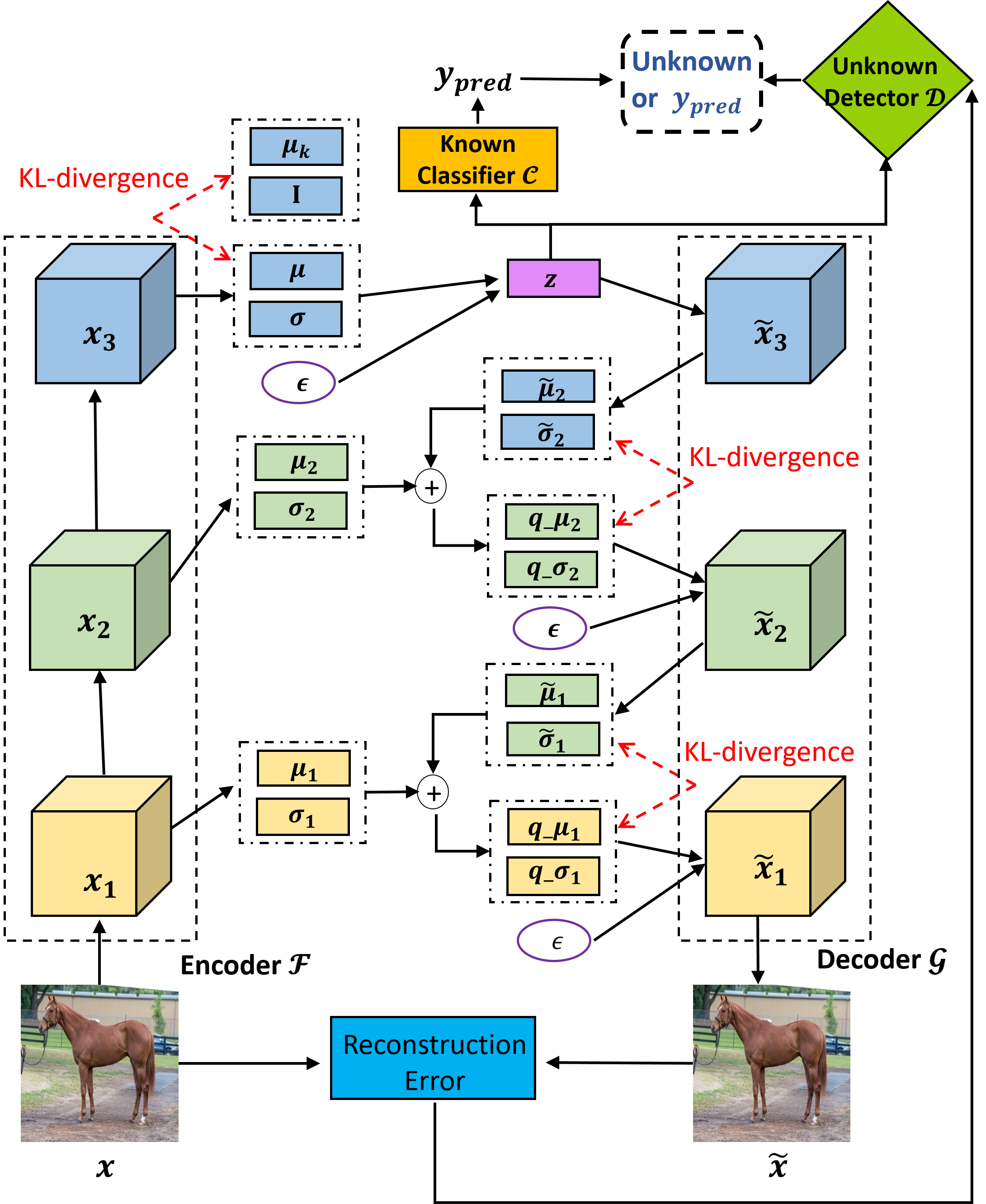}
    \caption{Block diagram of the proposed method: The \textbf{encoder} $\bm{{\mathcal{F}}}$ and \textbf{decoder} $\bm{{\mathcal{G}}}$ are applied with the probabilistic ladder architecture to extract high-level abstract latent features. The \textbf{known classifier} $\bm{{\mathcal{C}}}$ takes latent representations as input and produces the probability distribution over the known classes. The \textbf{unknown detector} $\bm{{\mathcal{D}}}$ is modeled by the conditional Gaussian distributions and reconstruction errors from training samples, which is used for unknown detection. During training, the proposed model is trained to minimize the sum of the reconstruction loss ${\mathcal{L}_r}$, KL-divergence ${\mathcal{L}_{KL}}$ (both in the latent space and middle layers) and classification loss ${\mathcal{L}_c}$. During testing, the \textbf{unknown detector} $\bm{{\mathcal{D}}}$ will judge whether this sampler is unknown by its latent features and reconstruction errors. If this sample is known, the \textbf{known classifier} $\bm{{\mathcal{C}}}$ will give out its predicted label.}  
    \label{fig:framework}
\end{figure}

The architecture of the proposed method is composed of four modules (as shown in Fig.~\ref{fig:framework}):\\
1. Encoder ${\mathcal{F}}$\\
2. Decoder ${\mathcal{G}}$\\
3. Known Classifier ${\mathcal{C}}$\\
4. Unknown Detector ${\mathcal{D}}$

\begin{figure} [htbp]
    \centering
    \includegraphics[width=\columnwidth]{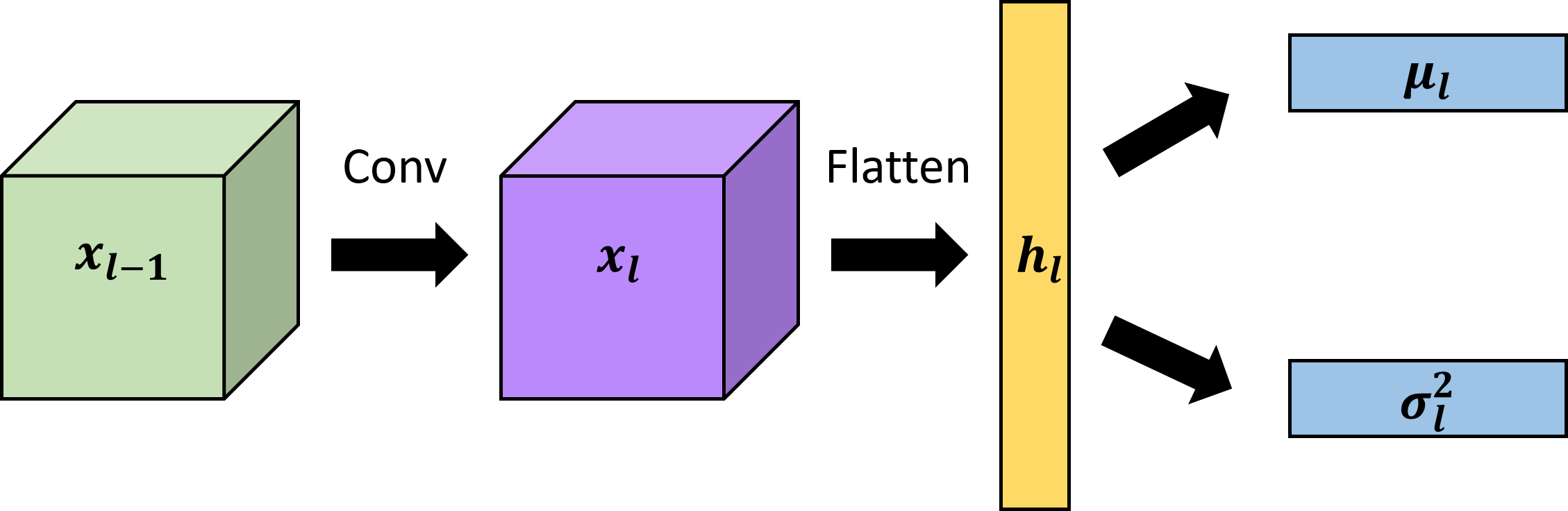}
    \caption{Operations in the upward pathway.}
    \label{fig:o1}
\end{figure}

\textbf{Encoder} $\bm{{\mathcal{F}}.}$ To extract high-level abstract latent features, the  probabilistic ladder architecture is adopted in each layer. In detail, the $l$-th layer in the encoder ${\mathcal{F}}$ is expressed as follows:

\begin{equation*}
\begin{aligned}
&\bm{x}_l=\text{Conv}(\bm{x}_{l-1})\\
&\bm{h}_l=\text{Flatten}(\bm{x}_l)\\
&\bm{\mu}_l=\text{Linear}(\bm{h}_l)\\
&\bm{\sigma}^{2}_l=\text{Softplus}(\text{Linear}(\bm{h}_l)
\end{aligned}
\end{equation*} 
where \verb Conv  is a convolutional layer followed by a batch-norm layer and a PReLU layer, \verb Flatten  is a linear layer to flatten 2-dimensional data into 1-dimension, \verb Linear  is a single linear layer and \verb Softplus  applies $\log(1+\text{exp}(\cdotp))$ non-linearity to each component of its argument vector (Fig.~\ref{fig:o1} illustrates these operations). The latent representation $\bm{z}$ is defined as $\bm{z}=\bm{\mu}+\bm{\sigma}\odot \bm{\epsilon}$ where $\bm{\epsilon}\sim {\mathcal{N}}(\bm{0},\textbf{I})$, $\odot$ is the element-wise product, and $\bm{\mu}$, $\bm{\sigma}$ are the outputs of the top layer $L$.

\begin{figure} [htbp]
    \centering
    \includegraphics[width=\columnwidth]{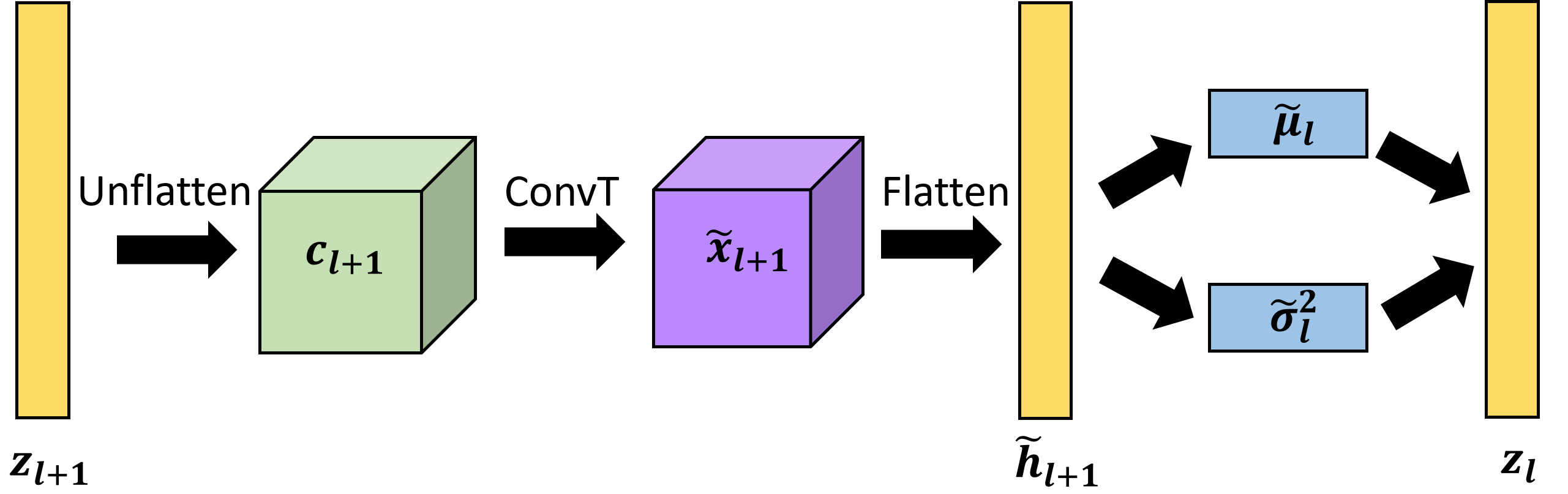}
    \caption{Operations in the downward pathway.}
    \label{fig:o2}
\end{figure}

\textbf{Decoder} $\bm{{\mathcal{G}}.}$ The $l$-th layer in the decoder ${\mathcal{G}}$ is expressed as follows:

\begin{equation*}
\begin{aligned}
&\bm{\tilde c}_{l+1}=\text{Unflatten}(\bm{\tilde z}_{l+1})\\
&\bm{\tilde x}_{l+1}=\text{ConvT}(\bm{\tilde c}_{l+1})\\
&\bm{\tilde h}_{l+1}=\text{Flatten}(\bm{\tilde x}_{l+1})\\
&\bm{\tilde \mu}_l=\text{Linear}(\bm{\tilde h}_{l+1})\\
&\bm{\tilde \sigma}^{2}_l=\text{Softplus}(\text{Linear}(\bm{\tilde h}_{l+1})\\
&\bm{z}_l=\bm{\tilde \mu}_l+\bm{\tilde \sigma}^{2}_l\odot \bm{\epsilon}
\end{aligned}
\end{equation*} 
where \verb ConvT  is a transposed convolutional layer and \verb Unflatten  is a linear layer to convert 1-dimensional data into 2-dimension (Fig.~\ref{fig:o2} illustrates these operations). In the $l$-th layer, the bottom-up information ($\bm{\mu}_l$ and $\bm{\sigma}_l$) and top-down information ($\bm{\tilde \mu}_l$ and $\bm{\tilde \sigma}_l$) are interacted by the following equations defined in \cite{ladder}:
\begin{equation}
\bm{q\_\mu}_{l}=\frac{\bm{\tilde \mu}_l\bm{\tilde \sigma}^{-2}_{l}+\bm{\mu}_l\bm{\sigma}^{-2}_{l}}{\bm{\tilde \sigma}^{-2}_{l}+\bm{\sigma}^{-2}_{l}}
\end{equation}
\begin{equation}
\bm{q\_\sigma^2}_{l}=\frac{1}{\bm{\tilde \sigma}^{-2}_{l}+\bm{\sigma}^{-2}_{l}}
\end{equation}

\textbf{Known Classifier} $\bm{{\mathcal{C}}.}$ The known classifier ${\mathcal{C}}$ is a Softmax layer, which takes the latent representation $\bm{z}$ as input. It .

\textbf{Unknown Detector} $\bm{{\mathcal{D}}.}$ When training is completed, the unknown detector ${\mathcal{D}}$ is modeled by information hidden in the latent representations and reconstruction errors. During the testing phase, the unknown detector ${\mathcal{D}}$ is used as a binary classifier to judge whether the input is known or unknown (details are discussed in Sec.~4.3).

\subsection{Training}
During the training phase, the proposed model forces the conditional posterior distributions $q_{\bm{\phi}}(\bm{z}|\bm{x},k)$ to approximate different multivariate Gaussian models $p_{\bm{\theta}}^{(k)}(\bm{z})={\mathcal{N}}(\bm{z};\bm{\mu}_{k},\textbf{I})$ where $k$ is the index of known classes, and the mean of $k$-th Gaussian distribution $\bm{\mu}_k$ is obtained by a fully-connected layer which maps the one-hot encoding of the input's label to the latent space. The KL-divergence in latent space (Eqn.~\ref{kl_vae}) is modified as follows:

\begin{equation}
\begin{aligned}
&\minus D_{KL}(q_{\bm{\phi}}(\bm{z}|\bm{x},k) \ || \ p_{\bm{\theta}}^{(k)}(\bm{z}))\\
&=\int q_{\bm{\phi}}(\bm{z}|\bm{x},k)\big(\log p_{\bm{\theta}}^{(k)}(\bm{z})-\log q_{\bm{\phi}}(\bm{z}|\bm{x},k)\big)d\bm{z}\\
&=\int {\mathcal{N}}(\bm{z};\bm{\mu},\bm{\sigma}^2)\big(\log {\mathcal{N}}(\bm{z};\bm{\mu}_{k},\textbf{I})-\log {\mathcal{N}}(\bm{z};\bm{\mu},\bm{\sigma}^2)\big)d\bm{z}\\
&=\frac{1}{2}\sum_{j=1}^{J}\big(1+\log(\sigma_{j}^{2})-(\mu_j-\mu_j^{(k)})^2-\sigma_{j}^{2}\big)
\end{aligned}
\end{equation}

During the training phase, the model is trained to minimize the sum of the reconstruction loss ${\mathcal{L}_r}$, KL-divergence ${\mathcal{L}_{KL}}$ and classification loss ${\mathcal{L}_c}$. To measure classification loss ${\mathcal{L}_c}$, we use softmax cross-entropy of prediction and ground-truth labels. To measure reconstruction loss ${\mathcal{L}_r}$, we use the ${{L}_1}$ distance between input images $\bm{x}$ and reconstructed image $\bm{\tilde x}$. As the  probabilistic ladder architecture is adopted, the KL-divergence is considered not only in the latent space but also in the middle layers:

\begin{equation}
\begin{aligned}
{\mathcal{L}_{KL}} = &\minus\frac{1}{L}\big[ D_{KL}\big(q_{\bm{\phi}}(\bm{z}|\bm{x},k) \ || \ p_{\bm{\theta}}^{(k)}(\bm{z})\big)\\
&+\sum_{l=1}^{L-1}D_{KL}\big(q_{\bm{\theta}}(\bm{\tilde x}_l|\bm{\tilde x}_{l+1}, \bm{x}) \ || \ q_{\bm{\theta}}(\bm{\tilde x}_l|\bm{\tilde x}_{l+1})\big)\big]
\label{kl}
\end{aligned}
\end{equation}
where 
\begin{equation}
q_{\bm{\theta}}(\bm{\tilde x}_l|\bm{\tilde x}_{l+1}, \bm{x})={\mathcal{N}}(\bm{\tilde x}_l;\bm{q\_\mu}_{l},\bm{q\_\sigma^2}_{l}) 
\end{equation}
\begin{equation}
q_{\bm{\theta}}(\bm{\tilde x}_l|\bm{\tilde x}_{l+1})={\mathcal{N}}(\bm{\tilde x}_l;\bm{\tilde \mu}_{l},\bm{\tilde \sigma^2}_{l})
\end{equation}

The loss function used in our model is summarized as follows:
\begin{equation}
{\mathcal{L}}=\minus({\mathcal{L}_r}+\beta{\mathcal{L}_{KL}}+\lambda{\mathcal{L}_{c})}
\label{loss}
\end{equation}
where $\beta$ is increased linearly from 0 to 1 during the training phase as described in \cite{ladder} and $\lambda$ is a constant. 

\subsection{Testing}
When training is completed, we model the per class multivariate Gaussian model $f_{k}(\bm{z})={\mathcal{N}}(\bm{z};\bm{m}_k,\bm{\sigma}_k^2)$ where $\bm{m}_k$ and $\bm{\sigma}_k^2$ are the mean and variance of the latent representations of all correctly classified training samples in $k$-th class. If the dimension of the latent space is $n$: $\bm{z}=(z_1,..., z_n)$, the probability of a sample locating in the distribution $f_{k}(\bm{z})$ is defined as follows:

\begin{equation}
P_k(\bm{z})=1-\int_{m_0-|z_0-m_0|}^{m_0+|z_0-m_0|} \cdots \int_{m_n-|z_n-m_n|}^{m_n+|z_n-m_n|} f_{k}(\bm{t}) \ d\bm{t}
\end{equation}

We also analyze information hidden in the reconstruction errors. The reconstruction errors of input from known classes are commonly smaller than that of unknown classes \cite{c2ae}. Here we obtain the reconstruction error threshold by ensuring 95\% training data to be recognized as known. Details of the testing procedure are described in Algo.~1. 

\begin{algorithm}
\caption{Testing procedure}\label{alg:euclid}
\begin{algorithmic}[1]
\Require Testing sample $\bm{X}$
\Require Trained modules ${\mathcal{F}}$, ${\mathcal{G}}$, ${\mathcal{C}}$
\Require Threshold $\tau_l$ of Gaussian distributions
\Require Threshold $\tau_r$ of reconstruction errors
\Require For each class $k$, let $\bm{z}_{i,k}$ is the latent representation of each correctly classified training sample $\bm{x}_{i,k}$
    \For {$k=1,\ldots,K$}
        \State compute the mean and variance of each class: $\bm{m}_k=mean_i(\bm{z}_{i,k})$, $\bm{\sigma}_k^2=var_i(\bm{z}_{i,k})$
        \State model the per class multivariate Gaussian: $f_{k}(\bm{z})={\mathcal{N}}(\bm{z};\bm{m}_k,\bm{\sigma}_k^2)$
    \EndFor
    \State latent representation $\bm{Z}={\mathcal{F}}(\bm{X})$
    \State predicted known label $y_{pred}=argmax({\mathcal{C}}\big(\bm{Z})\big)$
    \State reconstructed image $\bm{\tilde X}={\mathcal{G}}(\bm{Z})$
    \State reconstruction error $R=||\bm{X}-\tilde{\bm{X}}||_1$
    \If{${\forall}k\in \{1,...,K\}, P_k(\bm{Z})<\tau_l \ \textbf{or} \ R>\tau_r$ }
    \State predict $\bm{X}$ as unknown
    \Else
    \State predict $\bm{X}$ as known with label $y_{pred}$
    \EndIf
\end{algorithmic}
\end{algorithm}

\section{Experiments and Results}
\subsection{Implementation details}
In the proposed method, we use the SGD optimizer with a learning rate of 0.001, and fix the batch size to 64. The backbone is the re-designed VGGNet defined in \cite{crosr}. The dimensionality of the latent representation $\bm{z}$ is fixed to 32. For loss function described in Sec.~4.2, the parameter $\beta$ is increased linearly from 0 to 1 during the training phase as described in \cite{ladder}, while the parameter $\lambda$ is set equal to 100. The networks were trained without any large degradation in closed set accuracy from the original ones. The closed set accuracy of the networks for each dataset are listed in Table.~\ref{close}. The threshold $\tau_l$ of conditional Gaussian distributions is set to 0.5, and the threshold $\tau_r$ of reconstruction errors is obtained by ensuring 95\% training data be recognized as known. 

\subsection{Ablation Analysis}
In this section, we analyze our contributions from each component of the proposed method on CIFAR-100 dataset~\cite{cifar100}. The CIFAR-100 dataset consists of 100 classes, containing 500 training images and 100 testing images in each class. For ablation analysis, the performance is measured by F-measure (or F1-scores) \cite{f1} against varying Openness \cite{toward}. Openness is defined as follows:

\begin{equation}
    Openness=1-\sqrt{\frac {2\times N_{train}}{N_{test}+N_{target}}}
\end{equation}
where $N_{train}$ is the number of known classes seen during training, $N_{test}$ is the number of classes that will be observed during testing, and $N_{target}$ is the number of classes to be recognized during testing. We randomly sample 15 classes out of 100 classes as known classes and varying the number of unknown classes from 15 to 85, which means Openness is varied from 18\% to 49\%. The performance is evaluated by the macro-average F1-scores in 16 classes (15 known classes and \emph{unknown}).

\begin{table}
\begin{center}
\begin{tabular}{l c c c c}
\hline
Architecture  & MNIST & SVHN & CIFAR-10\\
\hline\hline
Plain CNN & 0.997 & 0.944 & 0.912\\
\hline
CGDL & 0.996 & 0.942 & 0.912\\
\hline
\end{tabular}
\end{center}
\caption{Comparison of closed set test accuracies between the plain CNN and the proposed method CGDL. Although the training objective of CGDL is classifying known samples as well as learning conditional Gaussian distributions, there is no significant degradation in closed set accuracy.}
\label{close}
\end{table}

We compare the following baselines for ablation analysis:\\
\textbf{I. CNN}: In this baseline, only the encoder ${\mathcal{F}}$ (without ladder architecture) and the known classifier ${\mathcal{C}}$ are trained for closed set classification. This model can be viewed as a traditional convolutional neural network (CNN). During testing, learned features will be fed to ${\mathcal{C}}$ to get the probability scores of known classes. A sample will be recognized as unknown if its probability score of predicted label is less than 0.5.\\
\textbf{II. CVAE}: The encoder ${\mathcal{F}}$, decoder ${\mathcal{G}}$ and classifier ${\mathcal{C}}$ are trained without the ladder architecture, and the testing procedure is the same as baseline I. This model can be viewed as a class conditional variational auto-encoder (CVAE).\\
\textbf{III. LCVAE}: The probabilistic ladder architecture is adopted in the CVAE, which contributes to the KL-divergences during training (Eqn.~\ref{kl}). We call this model as LCVAE. The testing procedure is the same as baseline I and II.\\
\textbf{IV. CVAE+CGD}: The model architecture and training procedure are the same as baseline II, while the conditional Gaussian distributions (CGD) are used to detect unknowns during testing.\\
\textbf{V. LCVAE+CGD}: In this baseline, LCVAE is introduced along with CGD-based unknown detector. The training and testing procedure are respectively the same as baseline III and IV.\\
\textbf{VI. LCVAE+RE}: Different from baseline V, reconstruction errors (RE), instead of CGD, are used in LCVAE to detect unknown samples.\\
\textbf{VII. Proposed Method}: The training procedure is the same as baseline V and VI, while during testing conditional Gaussian distributions and reconstruction errors are together used for unknown detection.\\

\begin{figure} [htbp]
    \centering
    \includegraphics[scale=0.38]{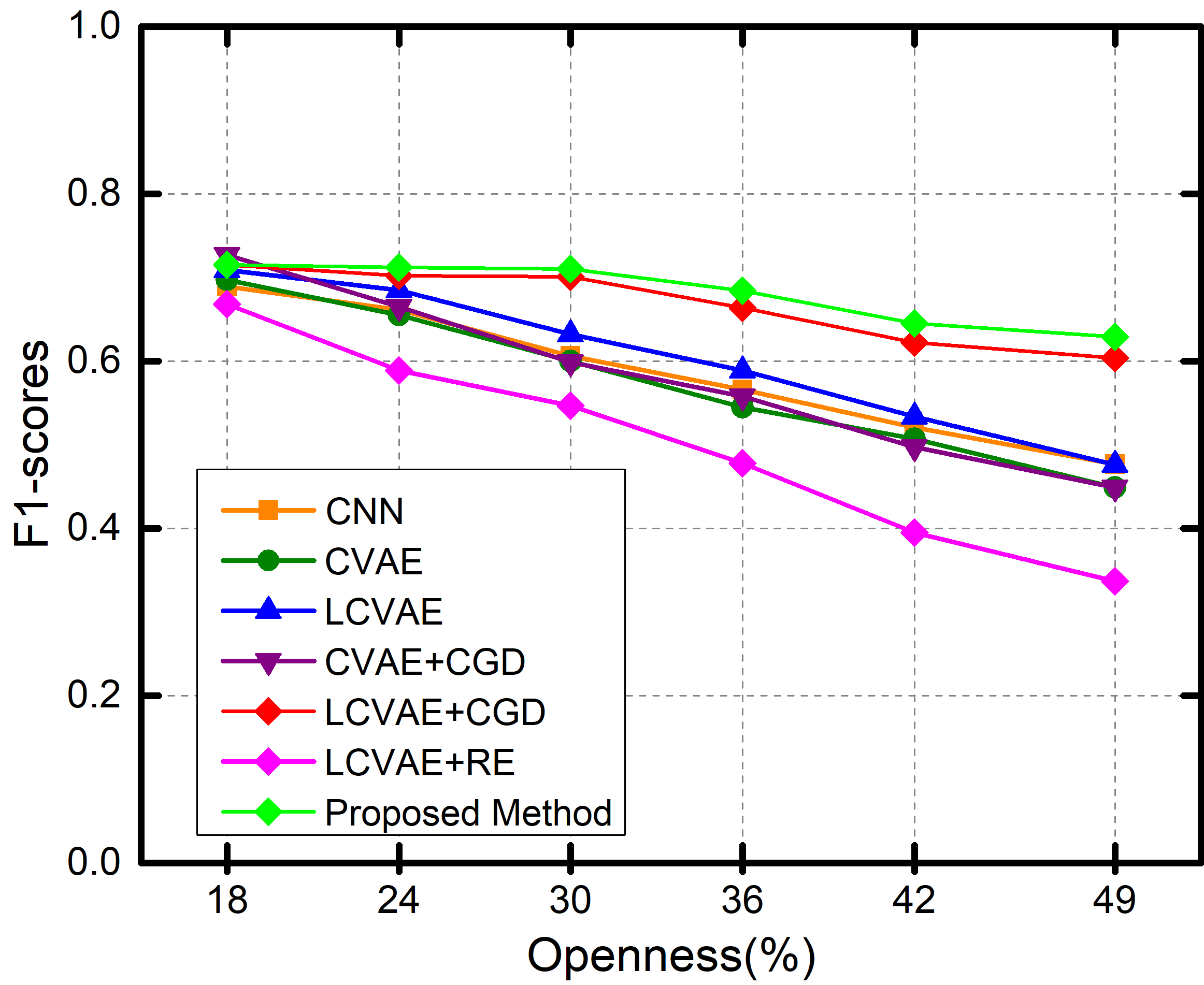}
    \caption{F1-scores against varying Openness with different baselines for ablation analysis.}
    \label{fig:abla}
\end{figure}

\begin{table*}
\begin{center}
\caption{Averaged F1-scores and their standard deviations in five randomized trials.}
\vspace{-2mm}
\label{state1}
\begin{tabular}{l c c c c c c}
\hline
Method & MNIST & SVHN & CIFAR10 & CIFAR+10 & CIFAR+50\\
\hline
\hline
Softmax & 0.768 $\pm$ 0.008 & 0.725 $\pm$ 0.012 & 0.600 $\pm$ 0.037 & 0.701 $\pm$ 0.012 & 0.637 $\pm$ 0.008 \\
Openmax~\cite{openmax} & 0.798 $\pm$ 0.018 & 0.737 $\pm$ 0.011 & 0.623 $\pm$ 0.038 & 0.731 $\pm$ 0.062 & 0.676 $\pm$ 0.056 \\
CROSR~\cite{crosr} & 0.803 $\pm$ 0.013 & 0.753 $\pm$ 0.019 & 0.668 $\pm$ 0.013 & 0.769 $\pm$ 0.016 & {0.684 $\pm$ 0.005} \\
GDFR~\cite{gdfr} & 0.821 $\pm$ 0.021 & 0.716 $\pm$ 0.010 & \textbf{0.700 $\pm$ 0.024} & \textbf{0.776 $\pm$ 0.003} & 0.683 $\pm$ 0.023 \\
\hline
CGDL (Ours) & \textbf{{0.837} $\pm$ {0.055}} & \textbf{{0.776} $\pm$ {0.040}} & {0.655} $\pm$ {0.023} & {0.760} $\pm$ {0.024} & \textbf{{0.695} $\pm$ {0.013}} \\
\hline
\end{tabular}
\end{center}
\end{table*}

The experimental results are shown in Fig.~\ref{fig:abla}. Among baseline I, II and III, unknown detection simply relies on the known classifier ${\mathcal{C}}$. Although the performance is a little improved when the probabilistic ladder architecture is adopted (baseline III), the overall performance in these three baselines is weak as the F1-scores degrade rapidly as the Openness increases. Conditional Gaussian distributions (CGD) is added for unknown detection in CVAE model (baseline IV), but it has seen no visible change in performance. In baseline V, this trend is alleviated by introducing CGD-based unknown detector into LCVAE. This shows the importance of the probabilistic ladder architecture for open set recognition. It is also the reason why the CGD-based unknown detection achieves better performance in baseline V than in baseline IV. If we only use reconstruction errors to detect unknowns (baseline VI), the results are worst. However, if reconstruction errors are added to the CGD-based unknown detector (baseline VII), there is a little improvement in performance. As a result, applying conditional Gaussian distributions and reconstruction errors to detect unknowns with the probabilistic ladder architecture achieves the best performance.

\subsection{Comparison with State-of-the-art Results}

\begin{table}
\begin{center}
\begin{tabular}{l c c c c}
\hline
Method & Omniglot & MNIST-noise & Noise \\
\hline\hline
Softmax & 0.595 & 0.801 & 0.829\\
Openmax \cite{openmax} & 0.780 & 0.816 & 0.826\\
CROSR \cite{crosr} & 0.793 & 0.827 & 0.826\\
\hline
ours: CGDL & \textbf{0.850} & \textbf{0.887} & \textbf{0.859}\\
\hline
\end{tabular}
\end{center}
\caption{Open set classification results on MNIST dataset with various outliers added to the test set as unknowns. The performance is evaluated by macro-averaged F1-scores in 11 classes (10 known classes and \emph{unknown}).}
\label{tab:state2}
\end{table}

In this section, we compare the proposed method with state-of-the-art methods,
and the OSR performance is measured by F1-scores in all known classes and \emph{unknown}. 

\begin{table*}
\begin{center}
\begin{tabular}{l c c c c}
\hline
Method & ImageNet-crop & ImageNet-resize & LSUN-crop & LSUN-resize \\
\hline\hline
Softmax \cite{crosr}$^*$ & 0.639 & 0.653 & 0.642 & 0.647 \\
Openmax \cite{openmax} & 0.660 & 0.684 & 0.657 & 0.668 \\
LadderNet+Softmax \cite{crosr} & 0.640 & 0.646 & 0.644 & 0.647 \\
LadderNet+Openmax \cite{crosr} & 0.653 & 0.670 & 0.652 & 0.659 \\
DHRNet+Softmax \cite{crosr} & 0.645 & 0.649 & 0.650 & 0.649 \\
DHRNet+Openmax \cite{crosr} & 0.655 & 0.675 & 0.656 & 0.664 \\
CROSR \cite{crosr} & 0.721 & 0.735 & 0.720 & 0.749\\
C2AE \cite{c2ae} & 0.837 & 0.826 & 0.783 & 0.801\\
\hline
ours: CGDL & \textbf{0.840} & \textbf{0.832} & \textbf{0.806} & \textbf{0.812}\\
\hline
\end{tabular}
\end{center}
\caption{Open set classification results on CIFAR-10 dataset with various outliers added to the test set as unknowns. The performance is evaluated by macro-averaged F1-scores in 11 classes (10 known classes and \emph{unknown}).\\ $^*$We report the experimental results reproduced in \cite{crosr}.}
\label{tab:state3}
\end{table*}

In Table~\ref{state1}, we conduct experiments on four standard image datasets: MNIST~\cite{mnist}, SVHN~\cite{svhn}, CIFAR10~\cite{cifar10} and CIFAR100~\cite{cifar100}. For each of the MNIST, SVHN, and CIFAR10 datasets, known classes consist of six randomly selected categories, and the remaining four categories constitute the unknown class. Meanwhile, in the CIFAR+10 /CIFAR+50 task, 4 vehicle categories in the CIFAR10 dataset (airplane, automobile, ship, and truck) are selected as known classes, and 10/50 animal categories in the CIFAR100 dataset are randomly selected to constitute the unknown class. The results shown in Table~\ref{state1} are averaged F1-scores and their standard deviations among 5 runs on different known and unknown separations. Noted that some previous works used Area under ROC Curve (AUROC) to measure the performance on these tasks, here we used released codes to evaluate OSR performance on macro-averaged F1-scores and report these results in Table~\ref{state1}. As shown in Table~\ref{state1}, on MNIST, SVHN and CIFAR+50 tasks, the proposed method achieves the highest accuracies.

\begin{figure} [htbp]
    \centering
    \includegraphics[scale=0.2]{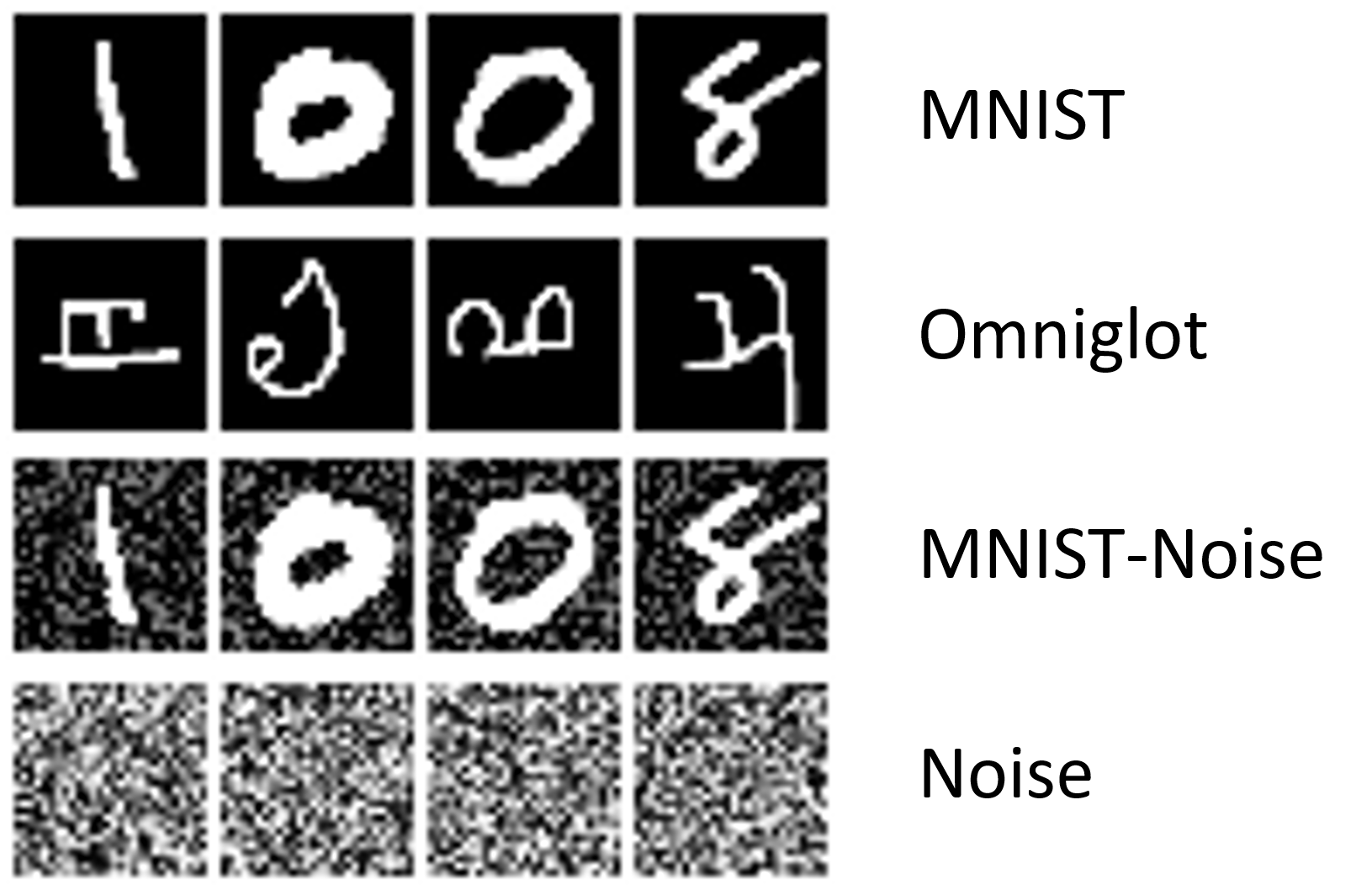}
    \caption{Examples from MNIST, Omniglot, MNIST-Noise, and Noise datasets.}
    \label{fig:out}
\end{figure}

We choose MNIST, the most popular hand-written digit dataset, as the training set. As outliers, we follow the set up in \cite{crosr}, using Omniglot \cite{omniglot}, MNIST-Noise, and Noise these three datasets. Omniglot is a dataset containing various alphabet characters. Noise is a synthesized dataset by setting each pixel value independently from a uniform distribution on $[0, 1]$. MNIST-Noise is also a synthesized dataset by adding noise on MNIST testing samples. Each dataset contains 10, 000 testing samples, the same as MNIST, and this makes the known-to-unknown ratio 1:1. Fig.~\ref{fig:out} shows examples of these datasets. The open set recognition scores are shown in Table.~\ref{tab:state2} and the proposed method achieves the best results on all given datasets.

Following the protocol defined in \cite{crosr}, all samples in CIFAR-10 dataset are collected as known data, and samples from other datasets, i.e., ImageNet \cite{imagenet} and LSUN \cite{lsun}, are selected as unknown samples. We resize or crop the unknown samples to make them have the same size with known samples. ImageNet-crop, ImageNet-resize, LSUN-crop, and LSUN-resize these four datasets are generated, and each dataset contains 10,000 testing images, which is the same as CIFAR-10. This makes during testing the known-to-unknown ratio 1:1. The performance of the method is evaluated by macro-averaged F1-scores in 11 classes (10 known classes and \emph{unknown}), and our results are shown in Table.~\ref{tab:state3}. We can see from the results that on all given datasets, the proposed method is more effective than previous methods and achieves a new state-of-the-art performance. 

\section{Conclusion}

In this paper, We have presented a novel method for open set recognition. Compared with previous methods solely based on VAEs, the proposed method can classify known samples as well as detect unknown samples by forcing posterior distributions in the latent space to approximate different Gaussian models. The probabilistic ladder architecture is adopted to preserve the information that may vanish in the middle layers. This ladder architecture obviously improves the open set performance. Moreover, reconstruction information is added to the unknown detector to further improve the performance. Experiments on several standard image datasets show that the proposed method significantly outperforms the baseline methods and achieves new state-of-the-art results.

{\small
\bibliographystyle{ieee_fullname}
\bibliography{egbib}
}

\end{document}